\documentclass[conference]{IEEEtran}
\usepackage{times}

\usepackage[numbers]{natbib}
\usepackage{multicol}
\usepackage[bookmarks=true]{hyperref}

\usepackage{wrapfig}
\usepackage{hyperref}
\usepackage{url}
\usepackage{graphicx}
\usepackage{amsmath}
\usepackage{amssymb}
\usepackage{nth}
\usepackage{booktabs}
\usepackage{times}
\usepackage{epsfig}
\usepackage{graphicx}
\usepackage{amsmath}
\usepackage{amssymb}
\usepackage{stfloats}
\usepackage{multirow}	
\usepackage{float}
\usepackage{xcolor}
\usepackage{booktabs}
\usepackage{multirow}
\usepackage{caption}
\usepackage{subcaption}
\usepackage{etoolbox}
\usepackage[normalem]{ulem}

\begin{document}

\title{Monocular Reconstruction of Neural Tactile Fields}


\author{Pavan Mantripragada, Siddhanth Deshmukh, Eadom Dessalene, Manas Desai, and Yiannis Aloimonos \\
Department of Computer Science, University of Maryland, College Park, USA\\
\texttt{Email: \{mppavan,iamsid,edessale,mdesai01,jyaloimo\}@umd.edu} \\
}


\maketitle


\begin{abstract}
Robots operating in the real world must plan through environments that deform, yield, and reconfigure under contact, requiring interaction-aware 3D representations that extend beyond static geometric occupancy. To address this, we introduce neural tactile fields, a novel 3D representation that maps spatial locations to the expected tactile response upon contact. Our model predicts these neural tactile fields from a single monocular RGB image -- the first method to do so. When integrated with off-the-shelf path planners, neural tactile fields enable robots to generate paths that avoid high-resistance objects while deliberately routing through low-resistance regions (e.g. foliage), rather than treating all occupied space as equally impassable. Empirically, our learning framework improves volumetric 3D reconstruction by $85.8\%$ and surface reconstruction by $26.7\%$ compared to state-of-the-art monocular 3D reconstruction methods (LRM and Direct3D).


\end{abstract}

\IEEEpeerreviewmaketitle

\section{Introduction}
\label{sec:introduction}

Robots increasingly operate in environments where objects are not fixed but move, yield, bend, or reconfigure under contact. Using vision to predict tactile outcomes is essential for performing contact-rich tasks in such environments. This requires more than perceiving where things are; robots must predict how objects will respond to applied force. To act safely and effectively, a robot must form an interaction-aware 3D representation \cite{bajcsy2018revisiting} of its surroundings -- one that encodes not only geometry, but also how the environment resists, absorbs, or transmits force.

Conventional methods for inferring 3D representations treat the world as static and non-interactive, producing reconstructions that describe only surface geometry \cite{3dconstructionmethods}. These static maps are commonly used as perceptual input for robots performing navigation or manipulation, but they provide no information about what happens when the robot actually makes contact. As a result, robots cannot reason about which regions are movable, compliant, or fragile, nor how much force should be applied during interaction.



\begin{figure}[t!] \centering \includegraphics[width=\columnwidth]{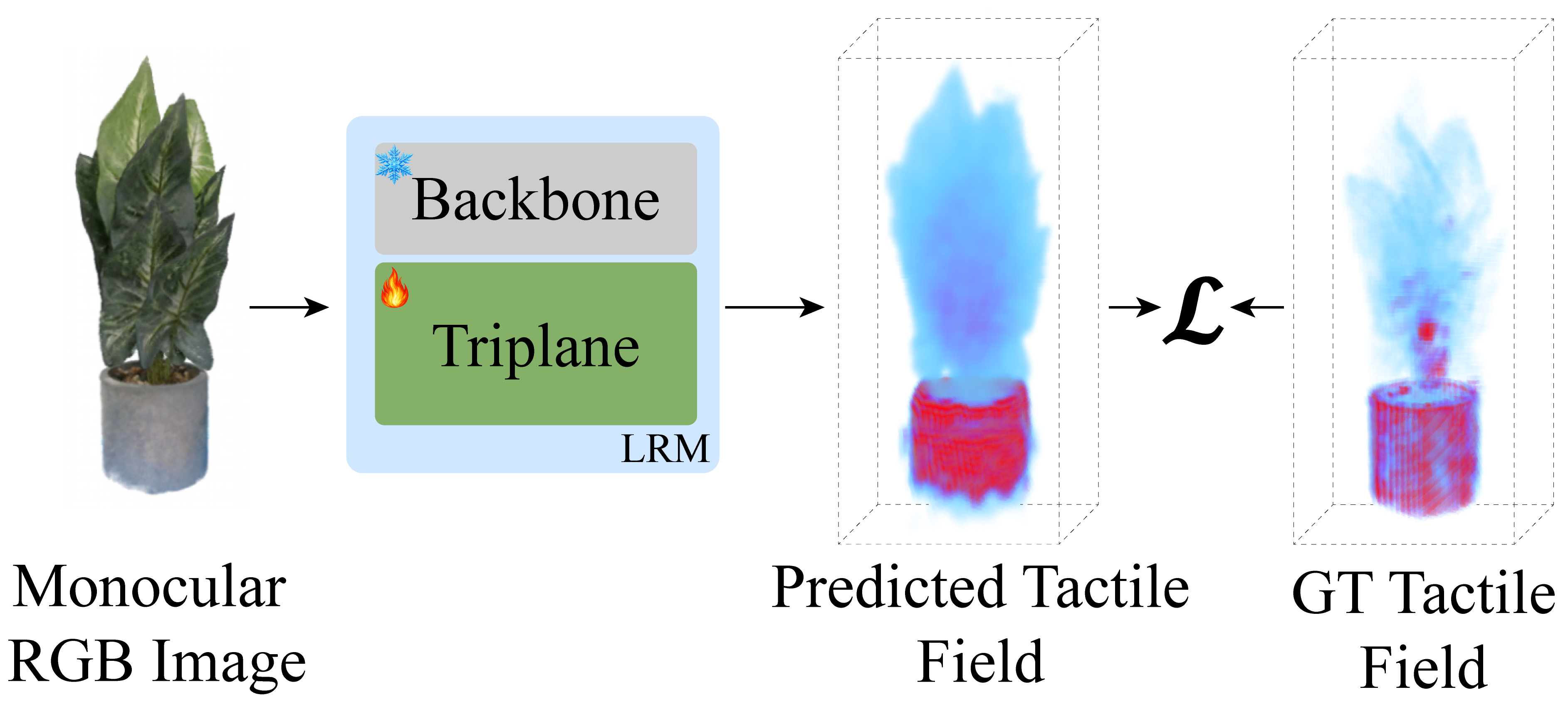} 
\caption{Our tactile field prediction pipeline. A monocular RGB image is processed through a Large Reconstruction Model (LRM) architecture with a frozen vision backbone (indicated by \includegraphics[height=1.5ex]{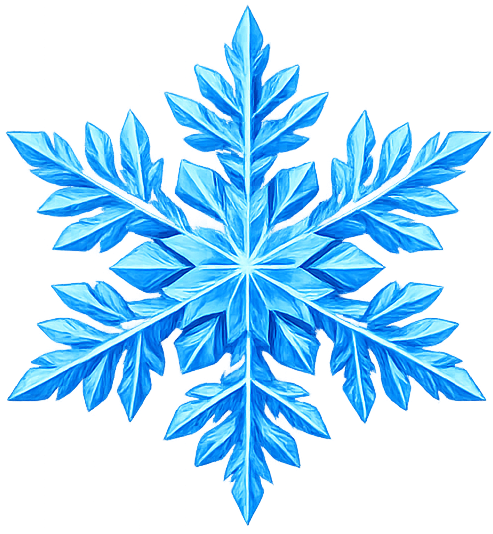}) and finetuned triplane decoder (indicated by \includegraphics[height=1.5ex]{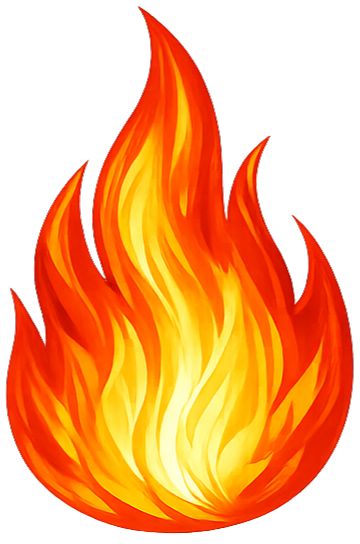}). The network outputs a continuous 3D tactile field prediction, which is supervised against the ground-truth neural tactile field constructed from physical interaction data (Section \ref{sec:datasetcollection}). Our model’s predictions encode the joint influence of object weight, compliance, and stability based solely on visual appearance.} 
\label{fig:network_prediction} 
\end{figure}

To address this limitation, we introduce \emph{3D neural tactile fields}, a representation that maps 3D positions to anticipated tactile perceptions. These neural tactile fields can be estimated from vision alone (see Figure \ref{fig:network_prediction} for an illustration), and the resulting prediction can be integrated with traditional planning and control frameworks. The values of the neural tactile fields emerge as a by-product of multiple physical properties of the contacted object: its weight, frictional characteristics, stability, and overall compliance. Consequently, neural tactile fields encode not only shape, but also the cost of physical interaction across the scene. Higher values correspond to regions exerting greater reactive forces, while lower values indicate compliant or easily displaced regions. 


\begin{figure}[t!] \centering \includegraphics[width=\columnwidth]{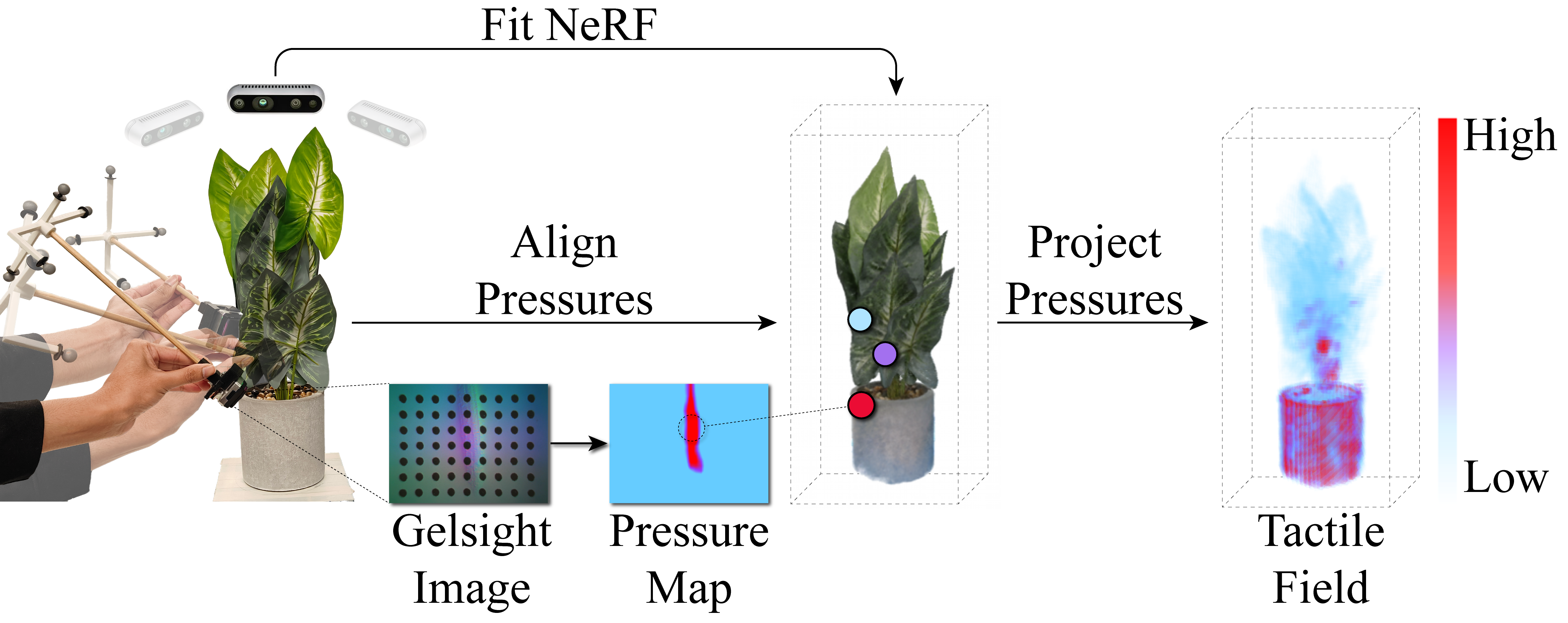} 
\caption{Ground-truth neural tactile field construction (dataset collection stage). Multi-view posed RGB images are used to fit a NeRF reconstruction of the object. A pose-tracked GelSight Mini sensor (shown contacting the plant) records pressure measurements at multiple surface locations. Each GelSight image is converted to a dense 2D pressure map over the sensor skin. These pressure measurements are then spatially aligned with the NeRF coordinate frame using the tracked sensor pose (``Align Pressures''). Finally, the aligned pressure points are projected into the 3D NeRF volume (``Project Pressures''), where each voxel aggregates nearby pressure values to produce a continuous 3D neural tactile field that serves as ground-truth supervision for training our prediction network.} 
\label{fig:groundtruth_generation} 
\end{figure}

To facilitate this, we collect and release a high-resolution tactile dataset containing both rigid and deformable objects, each grounded in accurate 3D reconstructions with dense tactile field supervision. Figure \ref{fig:groundtruth_generation} provides an illustration of the tactile field and how it is recorded. For each scene, a deformable tactile sensor is pressed against a set of points distributed across object surfaces, and the resulting pressure on the sensor during contact is recorded. Each data instance therefore provides (i) multiple calibrated RGBD viewpoints of the scene, (ii) a metric 3D NeRF reconstruction, and (iii) point-wise pressure measurements registered to that reconstruction. Our pressure measurements provide fine-grained physical supervision through direct physical interaction, an alternative to relying on manual image annotations that avoid direct interaction \cite{schwartz2019recognizing, makihara2022grasp}.


Using this collected data, we train a model to predict tactile fields directly from monocular RGB images. We build upon the Large Reconstruction Model (LRM) \cite{hong2023lrm}, a state-of-the-art single-view 3D reconstruction architecture. We finetune LRM on our tactile dataset, enabling the model to take a single RGB image as input and jointly predict geometry and tactile field values at each 3D location. To our knowledge, this is the first method to infer dense 3D tactile fields from monocular RGB input. Importantly, jointly optimizing our loss formulation over tactile fields not only enables tactile field prediction but also yields improved geometric reconstruction accuracy.

To further demonstrate the utility of our neural tactile fields, we integrate the predicted tactile fields into an off-the-shelf path planner tasked with solving minimum-cost path problems in two challenging scenarios. In the first, the planner must reach a goal while safely passing through deformable regions and avoiding rigid obstacles, illustrating contact-aware navigation through clutter. In the second, the planner must move lightweight, easily displaced objects aside while avoiding collisions with heavier, more stable ones. Across both settings, incorporating predicted tactile fields enables the planner to reason about traversable versus non-traversable obstacles.



The primary contributions of our work are as follows:

\begin{itemize}
    \item We introduce neural tactile fields -- a 3D representation that extends LRM to jointly reconstruct geometry and tactile field values from a monocular RGB image, enabling perception of physical properties without online physical interaction. We train a model over our dataset and plan to release the code weights upon acceptance.
    \item We integrate predicted tactile fields into path planning, demonstrating improved performance in settings that require reasoning about weight, deformability, and compliance.
    \item A high-resolution tactile dataset of rigid and deformable objects, with accurate 3D reconstructions and dense tactile field measurements across object volumes.
\end{itemize}

\section{Related Works}
\label{sec:related works}

In this section, we review prior work most relevant to our approach. Section \ref{subsec:single_view_recon} discusses methods that reconstruct scene from a monocular image. Section \ref{subsec:Tactile Sensing} surveys developments in tactile sensing and their use in robotic perception and manipulation. Section \ref{subsec:tactilereconstruction} examines approaches that combine visual and tactile modalities to infer physical properties and dynamics.

\subsection{Single-View 3D Reconstruction}
\label{subsec:single_view_recon}

Reconstructing objects and full scenes from a single image is a well-studied problem in computer vision. Popular approaches more recently reconstruct scenes as continuous 3D fields \cite{yu2021pixelnerf,xu2019disn,wang2023sparsenerf}. Reconstruction quality has been further improved through triplane-based architectures \cite{hong2023lrm,tochilkin2024triposr,zou2024triplane}, while generative modeling methods have enabled the synthesis of high-resolution 3D assets directly from images \cite{wu2024direct3d,li2025triposg}. However, these methods focus exclusively on geometric reconstruction and do not encode physical properties in their representations. In contrast, our approach reconstructs object geometry jointly with 3D physical properties from a single image.



\subsection{Tactile Sensing}
\label{subsec:Tactile Sensing}
Vision-based tactile sensors use internal cameras to observe the deformation of a soft medium, enabling high-resolution measurement of local surface geometry and contact forces. Representative designs include GelSight \cite{yuan2017gelsight, wang2021gelsight}, TacTip \cite{lepora2021pose}, and DIGIT \cite{lambeta2020digit}, each balancing optical fidelity and form factor. Such vision-based tactile sensing has proven crucial for fine-grained manipulation \cite{huang20243d, wang2022tacto}, contact localization \cite{kerr2022self, li2024biotactip}, and surface exploration \cite{shahidzadeh2024actexplore, de2021simultaneous}. Unlike these methods that directly use tactile input for control or state estimation, our work leverages tactile data purely as supervision for vision, enabling models to infer 3D, interaction-aware scene representations that encode how contact forces would distribute throughout the environment.

\subsection{Physically Grounded 3D Reconstruction with Tactile Sensing}
\label{subsec:tactilereconstruction}
Recent advances combine tactile feedback with visual sensing to augment 3D reconstruction, using contact information to refine geometry \cite{xu2023visual, zhao2023fingerslam}, estimate material properties \cite{lygerakis2025vitapes,guo2025robotic}, and build physically grounded representations \cite{stefani2025splattouch,comi2024touchsdf,dou2024tactile}. Tactile input has been fused with visual observations to improve surface completeness and reduce reconstruction uncertainty \cite{smith2021active, xu2023visual, zhao2023fingerslam, dave2024multimodal}, while other works learn continuous implicit shape representations directly from touch \cite{comi2024touchsdf, li2025tactile}. Active and exploratory touch strategies further target occluded or uncertain regions for data-efficient reconstruction \cite{smith2021active, wang2025touch2shape}. Collectively, these approaches demonstrate the potential of tactile sensing for 3D reconstruction. Our method differs from the majority of these approaches in that we employ tactile sensing only as offline supervision, enabling a vision-only model that predicts full 3D tactile fields encoding how the scene would respond to contact. The work most closely related to ours is \cite{han2025estimating}, which reconstructs a 3D neural stiffness field, a continuous volumetric representation of object stiffness, using multiple physical interactions collected with the Punyo bubble tactile sensor \cite{alspach2019soft}. In their extended work \cite{sbml}, stiffness is predicted over colored point clouds derived from RGB-D images, and is therefore limited to the observed portion of the object. In contrast, our approach predicts a similar but substantially simpler representation of the complete object directly from a single monocular RGB image. This enables a robot to reason about contact and plan appropriate interaction trajectories without requiring any prior physical interaction with the object.

\section{Method}
\label{sec:method}

In Section \ref{sec:datasetcollection}, we describe our data collection pipeline. We capture multi-view RGB images and fit a NeRF model for volumetric reconstruction \cite{mildenhall2021nerf}. We then collect tactile data with a GelSight Mini sensor, transform the pose of these measurements into the NeRF frame, and fuse them with the NeRF volume to obtain a continuous tactile field. In Section \ref{sec:tactilenetwork}, we introduce the single-view tactile field prediction network trained to regress average pressure at arbitrary 3D locations from a single RGB image input.

\subsection{Dataset Collection}
\label{sec:datasetcollection}
To train our model, we construct a paired visuotactile dataset that captures both the object's geometry and its physical response to contact. As illustrated in Figure \ref{fig:groundtruth_generation}, our collection process consists of two complementary components: multi-view RGB capture for NeRF-based visual reconstruction, and contact-based pressure measurements obtained with a GelSight Mini sensor. The following subsections detail each stage of this collection pipeline.

\subsubsection{Visual Reconstruction}
We first acquire a set of posed RGB images of the object using a pose-tracked Intel RealSense D435 camera~\cite{keselman2017intel}. Using this data, we fit a NeRF model around the object. The NeRF parameterization provides a continuous 5D vector field from which we can render photorealistic views and volumetric reconstructions of the object \cite{nerfstudio}. Following NeRF, we represent scene geometry using a density function $\sigma(\mathbf{x})$.

\begin{equation}
    \sigma(\mathbf{x}) : \mathbb{R}^3 \rightarrow \mathbb{R}_{\geq0}
\end{equation}

\subsubsection{Pressure Measurements}

While the NeRF reconstruction provides detailed scene geometry, it carries no information about how the object responds to contact. To obtain this complementary physical signal, we collect tactile pressure measurements using a GelSight Mini sensor \cite{yuan2017gelsight}. The pose of the sensor is tracked throughout interaction (Fig.~\ref{fig:groundtruth_generation}). During interaction, the sensor is pressed against the object at multiple locations. 

Each GelSight frame during these interactions is converted into a dense pressure map defined over a fixed 2D grid on the sensor skin using a pre-trained pressure estimation network \cite{helmut_dziarski2025feats}. Let
\begin{equation}
  \mathcal{G} = \{ \mathbf{u}_j \}_{j=1}^{J}, 
  \quad \mathbf{u}_j \in \mathbb{R}^2
\end{equation}
denote the discrete grid of skin coordinates on the elastomer surface. For each grid location $\mathbf{u}_j$, the network predicts a scalar pressure value $P_{j} \in \mathbb{R}_{\ge 0}$,
representing the pressure experienced locally at point $u_j$ on the sensor skin. We normalize pressures by the maximum measurable pressure $P_{\max}$ of the GelSight sensor before saturation:
\begin{equation}
  s_j = \frac{P_{j}}{P_{\max}}, 
  \qquad s_j \in [0,1].
\end{equation}.


Using the known pose of the sensor, we transform all skin points across time into the NeRF coordinate frame, producing a tactile point cloud $\mathcal{P}$ with corresponding tactile measurements.
\begin{equation}
  \mathcal{P} = \{ (\mathbf{p}_i, s_i) \}_{i=1}^N,
  \quad \mathbf{p}_i \in \mathbb{R}^3 .
\end{equation}

\subsubsection{Integration of Vision and Touch}
\label{subsubsec:integration_vision_touch}

With both the reconstructed visual geometry and the tactile point cloud in hand, we fuse the two modalities into a unified tactile field defined over the NeRF volume. The aggregated tactile point cloud approximately aligns with the reconstructed geometry.

For a query location $\mathbf{x} \in \mathbb{R}^3$, we aggregate nearby tactile measurements from the tactile point cloud $\mathcal{P}$. We consider all tactile samples within a Euclidean radius $r > 0$ of $\mathbf{x}$ and retain at most $K \in \mathbb{N}$ samples with the largest normalized pressure values. We denote this selected set by $\mathcal{N}_K(\mathbf{x})$:
\begin{equation}
  \mathcal{N}_K(\mathbf{x}) =
  \arg\max_K
  \left\{ s_i \;\middle|\;
  (\mathbf{p}_i,s_i)\in\mathcal{P},\;
  \|\mathbf{p}_i-\mathbf{x}\|_2 \le r
  \right\}.
\end{equation}

We define our neural tactile field
$S : \mathbb{R}^3 \rightarrow \mathbb{R}_{\ge 0}$,
where $S(\mathbf{x})$ represents the aggregated normalized pressure at location $\mathbf{x}$.
\begin{equation}
  S(\mathbf{x}) =
  \begin{cases}
    \dfrac{1}{|\mathcal{N}_K(\mathbf{x})|}
    \sum\limits_{i\in\mathcal{N}_K(\mathbf{x})} s_i,
    & \sigma(\mathbf{x})>0,\; |\mathcal{N}_K(\mathbf{x})|>0, \\[0.6em]
    \varepsilon,
    & \sigma(\mathbf{x})>0,\; |\mathcal{N}_K(\mathbf{x})|=0, \\[0.4em]
    0, & \text{otherwise},
  \end{cases}
  \label{S_eqn}
\end{equation}
where $\varepsilon$ is a small positive constant.

We rescale and transform the coordinate frame so that the tactile field $S(x)$ is centered at the centroid of the bounding box of the object and contained within the $[-1,1]^3$ cube. We then voxelize the tactile field onto a $100\times100\times100$ voxel grid, which serves as supervision for the prediction network $f_\theta$.


\subsubsection{Dataset Details}
\begin{figure}[b!] \centering \includegraphics[width=\columnwidth]{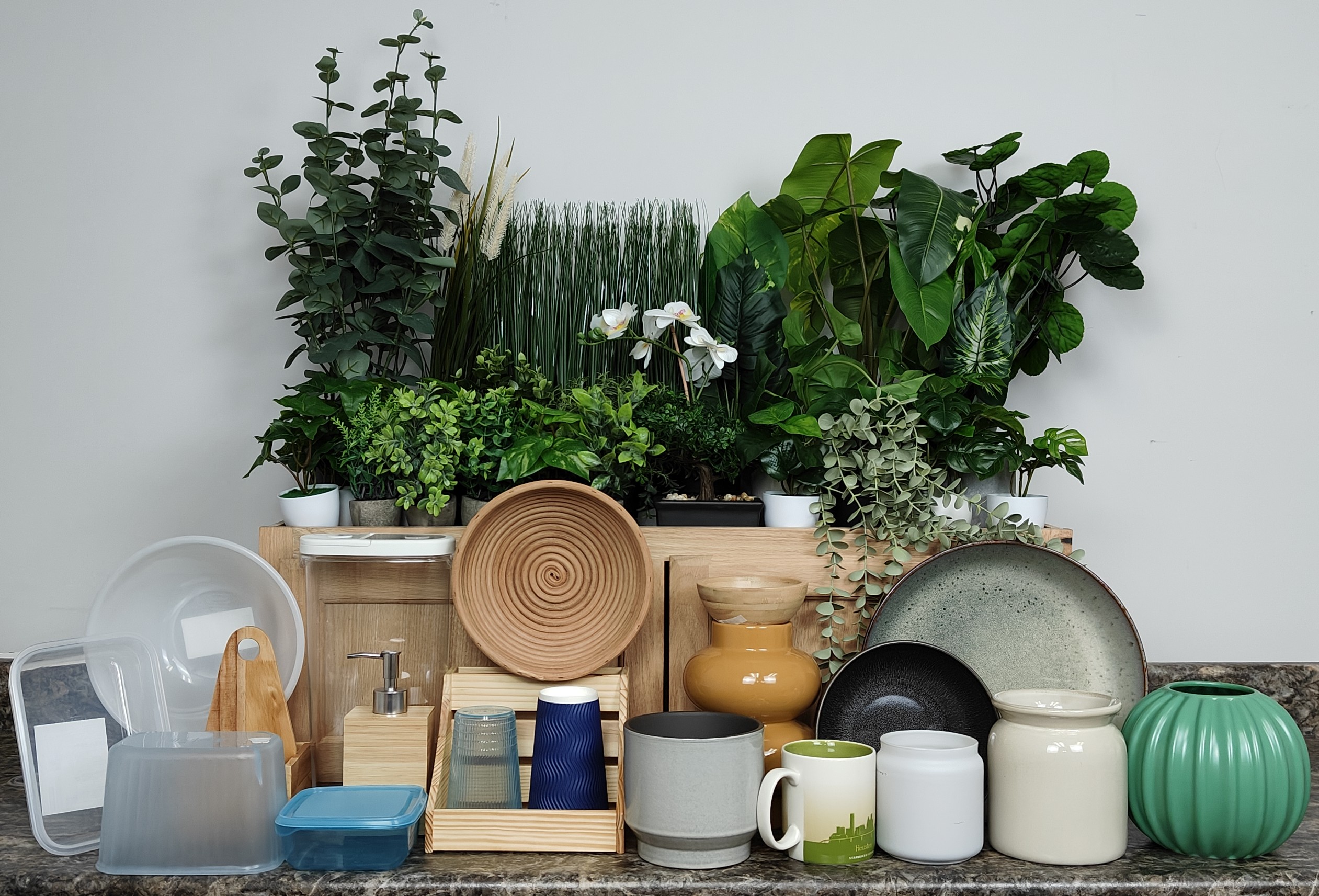} 
\caption{The 40 objects used for training and evaluation, spanning a range of weights, stiffnesses, and materials. Plant pots were rigidly mounted during data collection, while all other objects were free to move under contact.} 
\label{fig:dataset} 
\end{figure}

The dataset comprises $20$ plants and $20$ household objects of varying weights. See Figure \ref{fig:dataset} for the objects used. For each object, approximately $70$ images were captured, with the camera pose recorded for every image. Corresponding tactile data was collected over $7,000$ to $10,000$ tactile frames depending on the size of the object. When pressing the objects with the sensor, the plants were fixed rigidly whereas the remaining objects were allowed to move slightly.  For model training and evaluation, $32$ objects were used for training and $8$ objects were held out for validation. 

To enable pose tracking, the camera and the tactile sensor were equipped with four reflective tracking markers each. Ground-truth sensor poses were obtained using an OptiTrack motion capture system consisting of eight infrared cameras. The system provided sensor poses in a global coordinate frame, along with timestamps recorded during data acquisition. These timestamps were later used to temporally align and fuse the sensor poses with the GelSight measurements, providing ground-truth pose annotations.

\subsection{Tactile Field Prediction Network}
\label{sec:tactilenetwork}

We train a Large Reconstruction Model (LRM) \cite{hong2023lrm} as the backbone for our tactile field prediction network $f_\theta$ (Figure \ref{fig:network_prediction}). We freeze the vision backbone and image-to-triplane decoder while finetuning the triplane upsampler and decoder on our tactile dataset. Our key insight is to repurpose LRM's density predictions as tactile field values. 

Given an RGB image $I$, the original LRM network \cite{hong2023lrm} $f_\theta$ predicts two outputs at any 3D location $\mathbf{x}$ and viewing direction $\mathbf{d}$: the RGB color $\mathbf{c}$ and density $\sigma$:

\begin{equation}
\mathbf{c}, \sigma = f_{\theta}(\mathbf{x}, \mathbf{d}, I)
\end{equation}



We define our predicted tactile field $\hat{S}(\mathbf{x})$ as the scaled density output:

\begin{equation}
\hat{S}(\mathbf{x}) = \frac{\sigma(\mathbf{x})}{\gamma}
\end{equation}

where $\gamma > 0$ is a scaling factor that maps density values to the pressure range [0, 1]. 

For rendering purposes, we use the scaled tactile field prediction $\gamma \hat{S}(\mathbf{x})$ in place of density in the standard volumetric rendering equation. This allows us to supervise the tactile field predictions $\hat{S}$ directly with $S$ while maintaining compatibility with LRM's photometric rendering loss using $\gamma \hat{S}$.



\subsubsection{Loss}

Let $S$ denote the ground-truth neural tactile field from Equation \ref{S_eqn}. We train the network $f_{\theta}$ using two complementary losses: a volumetric reconstruction loss $\mathcal{L}_v$ that supervises the predicted tactile field $\hat{S}$ against the ground-truth $S$, and a photometric reconstruction loss $\mathcal{L}_{p}$ as defined in \cite{hong2023lrm} that supervises the color predictions $\mathbf{c}$ and ensures geometric consistency through $\gamma \hat{S}$.

The volumetric loss $\mathcal{L}_v$ provides direct 3D supervision and consists of two components:

\begin{equation}
    \mathcal{L}_{v} = \lambda_1 \mathcal{L}_{wMSE} + \lambda_2\mathcal{L}_{IoU}
\end{equation}

The first component, $\mathcal{L}_{wMSE}$, computes a weighted mean squared error between $\hat{S}_i$, the value of the predicted tactile field $\hat{S}$ at location $i$, and the ground-truth tactile field voxel $S_i$ obtained as described in Section \ref{subsubsec:integration_vision_touch}. We employ weighted loss to handle the class imbalance between contact and non-contact regions. For all prediction-target pairs $(\hat{S}_i, S_i)$, we define the weighted MSE loss as
\begin{equation}
  \mathcal{L}_{wMSE} =
  \begin{cases}
    \frac{w_1}{A}\sum_{i=1}^A (\hat{S}_i - S_i)^2,
      & \text{if } S_i > \epsilon, \\[0.4em]
    \frac{w_2}{B}\sum_{i=1}^B (\hat{S}_i - S_i)^2,
      & \text{otherwise,}
  \end{cases}
  \label{eq:weighted-mse}
\end{equation}
where $\epsilon > 0$ is a small threshold that distinguishes contact from non-contact regions, where $A$ and $B$ are the number of samples denoting contact and no-contact regions, and $w_1 > w_2 > 0$ are positive weights that upweight errors in regions with significant tactile interaction. This formulation prevents training from being dominated by the many zero-valued samples in the sparse tactile field.

Since MSE loss is highly sensitive to translational and rotational offsets of the predicted field relative to the ground-truth, we add a second component: a differentiable approximation of Intersection-over-Union as a loss term $\mathcal{L}_{IoU}$ between $\hat{S}$ and $S$:


\begin{equation}
\begin{gathered}
\hat{O}_{ij} = \text{sigmoid}(\beta(\hat{S}_i-\tau_j)) \\
O_{ij} = \text{sigmoid}(\beta(S_i-\tau_j)) \\
\mathcal{L}_{IoU} = 1-
\frac{1}{8} \sum_{j=1}^8
\frac{\sum_{i=1}^{100^3} \hat{O}_{ij} \cdot O_{ij}}
     {\sum_{i=1}^{100^3} (\hat{O}_{ij}+O_{ij}-\hat{O}_{ij}\cdot O_{ij})}
\end{gathered}
\end{equation}

where $\hat{O}_{ij}$ and $O_{ij}$ represent predicted and ground-truth soft occupancies of the $\tau_j$ superlevel-set of the tactile field, $\beta$ is a positive constant that controls the sharpness of the soft occupancy. The outer sum is computed over $8$ superlevel sets, and the inner sums are computed over $100^3$ voxels.

Combining the photometric and volumetric losses, the final training objective is:

\begin{equation}
    \mathcal{L} = \lambda_3 \mathcal{L}_{p} + \lambda_4 \mathcal{L}_{v}
\end{equation}
where $\lambda_3$ and $\lambda_4$ are loss weights that balance photometric consistency with direct tactile field supervision.

\subsubsection{Training Details}
\label{subsec:training_details}
We train the model using the AdamW optimizer with a cosine annealing learning rate schedule. The learning rate is initialized at $5\times10^{-3}$ and decayed to $1\times10^{-4}$, with optimizer parameters $\beta_1=0.9$ and $\beta_2=0.95$. Training is performed for 50 epochs with a batch size of 128, distributed across 64 NVIDIA RTX A4000 GPUs.

For the photometric reconstruction loss, we follow the LRM training protocol and render randomly sampled $128\times128$ image patches from three views: the input view and two randomly selected side views. This encourages geometric consistency while maintaining computational efficiency.

The weighted MSE loss uses weights $w_1=0.8$ and $w_2=0.2$ to emphasize regions with tactile contact. The volumetric loss weights are set to $\lambda_1=0.5$ for $\mathcal{L}_{wMSE}$ and $\lambda_2=1.0$ for $\mathcal{L}_{IoU}$. The final objective balances photometric and volumetric supervision with $\lambda_3=0.25$ and $\lambda_4=0.5$, respectively. We fix the density-to-tactile scaling factor to $\gamma=100$ for all experiments.

\section{Experiments}
\label{sec:experiments}


We first define a key concept used throughout our evaluation. An interaction threshold $\tau \in \mathbb{R}_{\ge 0}$ is used to extract superlevel-sets from the tactile field,  enabling us to evaluate reconstruction quality at different pressure levels and to define collision spaces for planning.

Let $\mathcal{X}$ denote a regular voxel grid inside the cube $[-1,1]^3$. For a given threshold $\tau$, we define the predicted and ground-truth superlevel sets as 
\begin{equation}
  \hat{\mathcal{M}}_{\tau} =
  \left\{ \mathbf{x} \in \mathcal{X} \;\middle|\; \hat{S}(\mathbf{x}) > \tau \right\}
  \label{eq:mpred}
\end{equation}
\begin{equation}
  \mathcal{M}_{\tau} =
  \left\{ \mathbf{x} \in \mathcal{X} \;\middle|\; S(\mathbf{x}) > \tau \right\}
  \label{eq:mgt}
\end{equation}
and the surfaces of these superlevel-sets as $\hat{\mathcal{S}}_{\tau}$ and $\mathcal{S}_{\tau}$ respectively. Intuitively, lower values of $\tau$ select the entire object's volume as the superlevel set, while higher values of $\tau$ focus only on regions with high tactile pressure during interaction.

Section \ref{subsec:monocular 3D reconstructions} evaluates how closely predicted tactile fields match ground-truth tactile fields. Section \ref{subsec:interaction aware planning} assesses whether predicted tactile fields support more efficient, interaction-aware planning compared to geometry-only planners.





\subsection{Monocular 3D Reconstruction}
\label{subsec:monocular 3D reconstructions}

We evaluate our tactile field prediction network on $8$ unseen objects, over $120$ rendered views per object. Given an RGB image of the object, the network predicts a continuous tactile field, which we compare against the ground-truth neural tactile field. We evaluate reconstruction quality across multiple values of $\tau$ using three standard metrics: Intersection-over-Union (IoU), Chamfer Distance (CD), and F1 score, as shown in Figures  \ref{fig:recon_chamfer}, \ref{fig:recon_f1}, and \ref{fig:recon_iou}.


\subsubsection{Intersection-over-Union (IoU)}
\label{subsubsec:IoU_metric}
Following the IoU definition in \cite{mescheder2019occupancy}, we define the IoU at threshold $\tau$ as
\begin{equation}
  \mathrm{IoU}_{\tau}
  =
  \frac{\left| \hat{\mathcal{M}}_{\tau} \cap \mathcal{M}_{\tau} \right|}
       {\left| \hat{\mathcal{M}}_{\tau} \cup \mathcal{M}_{\tau} \right|}
  \label{eq:IoU}
\end{equation}
where $|\cdot|$ denotes the cardinality of the corresponding set. This metric measures volumetric overlap between predicted and ground-truth superlevel sets.

\subsubsection{Chamfer Distance (CD)}
\label{subsubsec:CD_metric}
To evaluate surface reconstruction quality, we compute the bidirectional Chamfer Distance between predicted and ground-truth superlevel-set surfaces:

\begin{align}
\mathrm{CD}^{(1)}_{\tau}
&=
\frac{1}{|\hat{\mathcal S}_{\tau}|}
\sum_{\hat{\mathbf x}\in\hat{\mathcal S}_{\tau}}
\min_{\mathbf y\in\mathcal S_{\tau}}
\|\hat{\mathbf x}-\mathbf y\|_2, \\[0.4em]
\mathrm{CD}^{(2)}_{\tau}
&=
\frac{1}{|\mathcal S_{\tau}|}
\sum_{\mathbf y\in\mathcal S_{\tau}}
\min_{\hat{\mathbf x}\in\hat{\mathcal S}_{\tau}}
\|\mathbf y-\hat{\mathbf x}\|_2 .
\end{align}

\begin{equation}
\mathrm{CD}_{\tau} =
\begin{cases}
\mathrm{CD}^{(1)}_{\tau} + \mathrm{CD}^{(2)}_{\tau},
& |\hat{\mathcal S}_{\tau}|>0,\ |\mathcal S_{\tau}|>0, \\[0.6em]
1.0,
& |\hat{\mathcal S}_{\tau}|>0,\ |\mathcal S_{\tau}|=0, \\[0.4em]
1.0,
& |\hat{\mathcal S}_{\tau}|=0,\ |\mathcal S_{\tau}|>0, \\[0.4em]
0,
& \text{otherwise},
\end{cases}
\label{eq:CD}
\end{equation}

Lower Chamfer Distance indicates better surface alignment between prediction and ground truth.

\subsubsection{F1 Score}
\label{subsubsec:F1score_metric}
We also compute precision and recall at a distance threshold of $0.05$, then combine them into an F1 score:

\begin{align}
\mathrm{P}_{\tau}
&=
\frac{1}{|\hat{\mathcal S}_{\tau}|}
\sum_{\hat{\mathbf x}\in\hat{\mathcal S}_{\tau}}
\mathbb{I}\!\left[
\min_{\mathbf y\in\mathcal S_{\tau}}
\|\hat{\mathbf x}-\mathbf y\|_2 \le 0.05
\right], \\[0.4em]
\mathrm{R}_{\tau}
&=
\frac{1}{|\mathcal S_{\tau}|}
\sum_{\mathbf y\in\mathcal S_{\tau}}
\mathbb{I}\!\left[
\min_{\hat{\mathbf x}\in\hat{\mathcal S}_{\tau}}
\|\mathbf y-\hat{\mathbf x}\|_2 \le 0.05
\right],
\end{align}
where $\mathbb{I}$ represents an indicator function.

\begin{equation}
\mathrm{F1}_{\tau} =
\begin{cases}
\displaystyle
\frac{2\,\mathrm{P}_{\tau}\,\mathrm{R}_{\tau}}
{\mathrm{P}_{\tau}+\mathrm{R}_{\tau}},
& |\hat{\mathcal S}_{\tau}|>0,\ |\mathcal S_{\tau}|>0, \\[0.6em]
0, & |\hat{\mathcal S}_{\tau}|>0,\ |\mathcal S_{\tau}|=0, \\[0.4em]
0, & |\hat{\mathcal S}_{\tau}|=0,\ |\mathcal S_{\tau}|>0, \\[0.4em]
1, & \text{otherwise},
\end{cases}
\label{eq:F1}
\end{equation}

The F1 score provides a balanced measure of reconstruction accuracy that accounts for both false positives and false negatives.

\subsection{Interaction Aware Planning}
\label{subsec:interaction aware planning}
To evaluate the benefits of interaction-aware tactile fields in path planning, we consider the problem of finding shortest paths that never exceed a prescribed interaction threshold $\tau$.

Given an interaction threshold $\tau$, we treat $\hat{\mathcal M}_{\tau}$ as the collision-space of the planner. For each start–goal pair, we then seek the shortest path that avoids $\hat{\mathcal M}_{\tau}$.

We conduct $T$ planning trials using an RRT* planner~\cite{rrtstar}. For each trial, we randomly sample a start and goal position uniformly on a circle of radius $1.5$ centered at the origin, at heights $0.0$ and $-0.8$ for movable objects and plants, respectively. For each start-goal pair, the planner returns a collision-free path with length $L_{\tau}^{(t)}$ for trial $t$.


\begin{figure}[t]
    \centering
    \includegraphics[width=.95\linewidth]{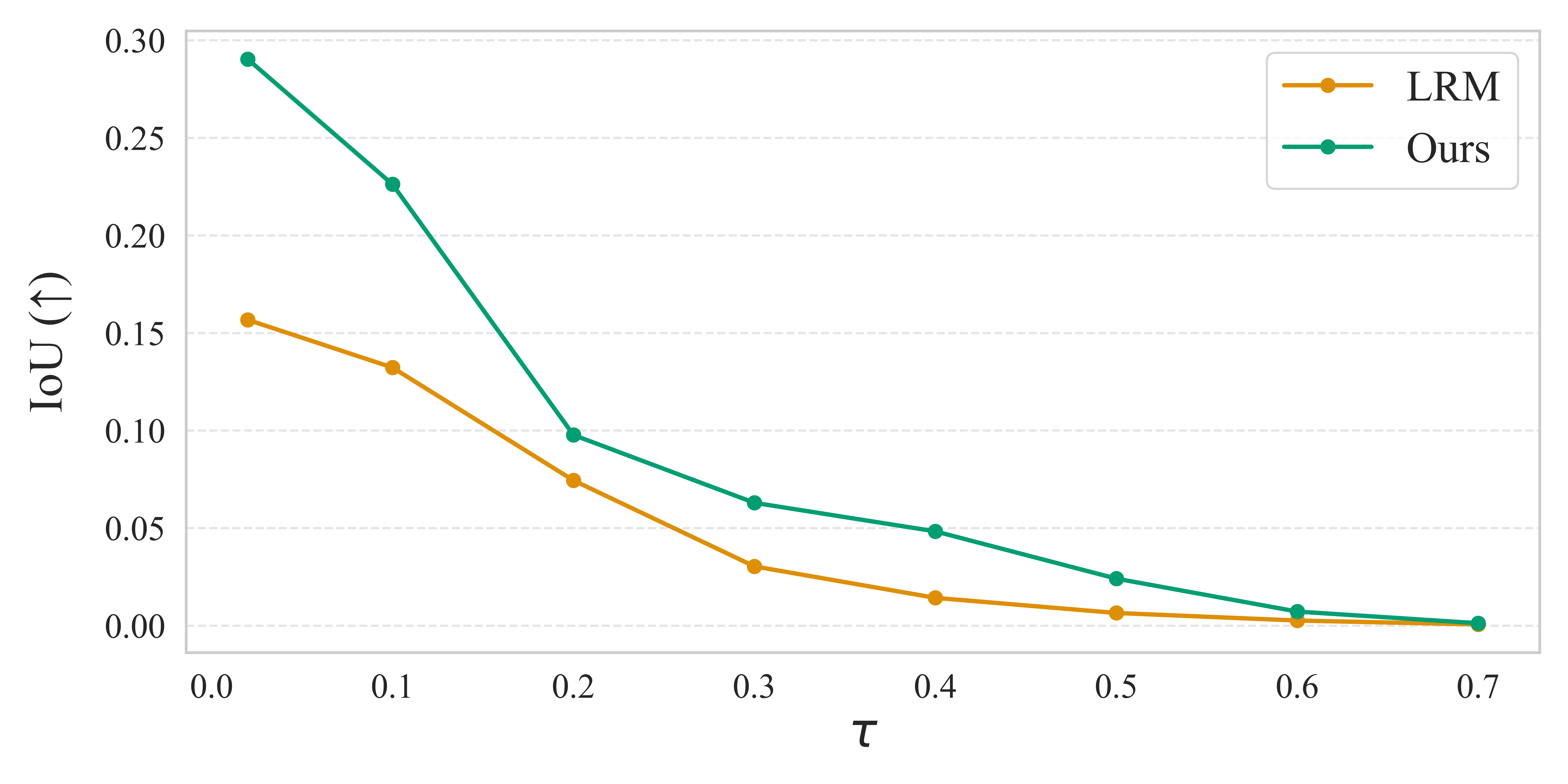}
    \caption{\textbf{Intersection-over-Union (Higher $\uparrow$ is better)} for unseen object reconstruction as a function of the interaction threshold $\tau$. Our method consistently outperforms LRM across all $\tau$ values. We do not include Direct3D as a comparison as their predicted meshes are not view-aligned. }
    \label{fig:recon_iou}
\end{figure}

\begin{figure}[!t]
    \centering
    \includegraphics[width=.95\linewidth]{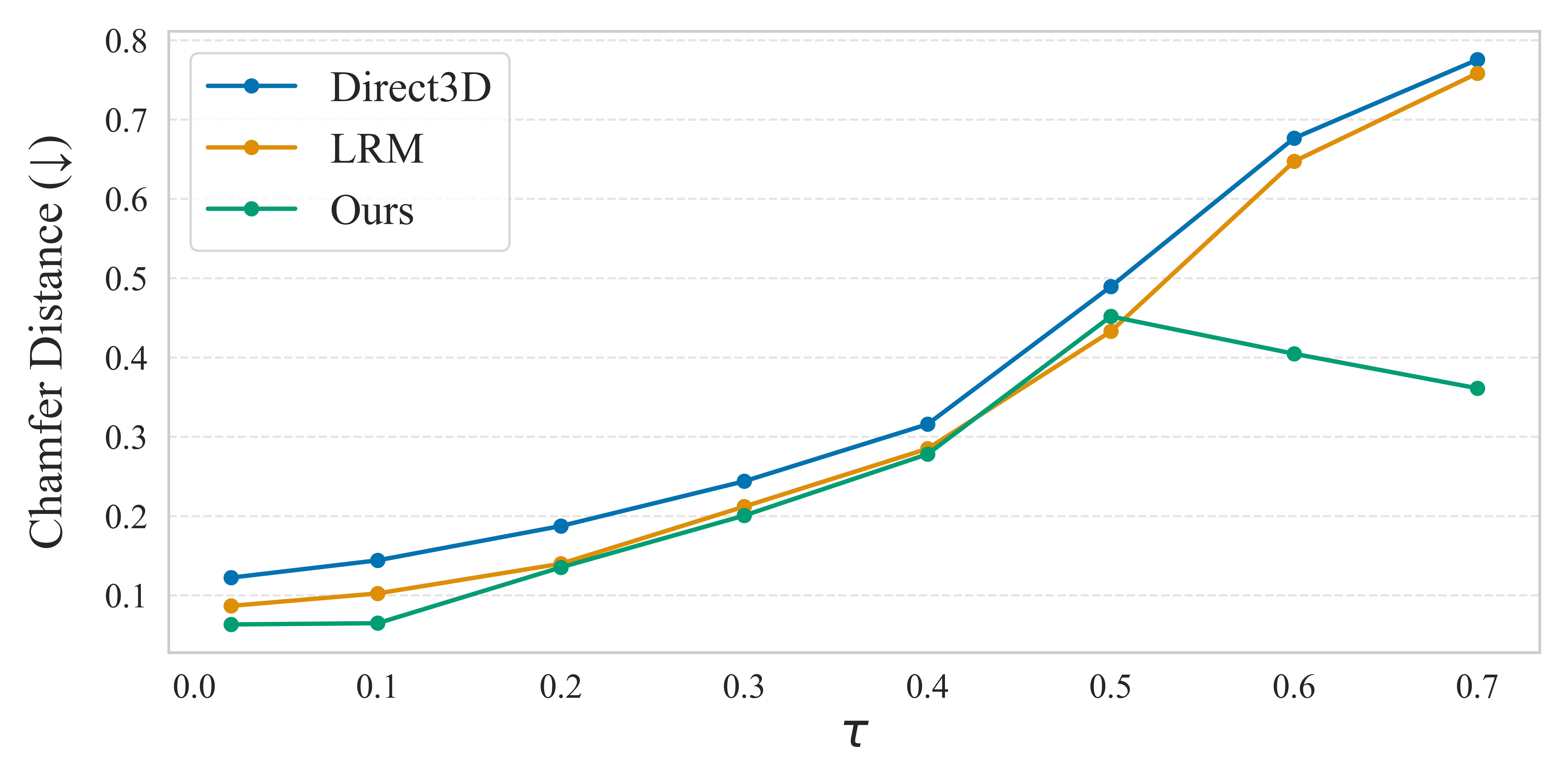}
    \caption{\textbf{Chamfer Distance (Lower $\downarrow$ is better)} for unseen object reconstruction as a function of the interaction threshold $\tau$. Lower values indicate more accurate surface reconstruction, with our method achieving consistently lower error than LRM and Direct3D across all thresholds.}
    \label{fig:recon_chamfer}
\end{figure}

\begin{figure}[t]
    \centering
    \includegraphics[width=.95\linewidth]{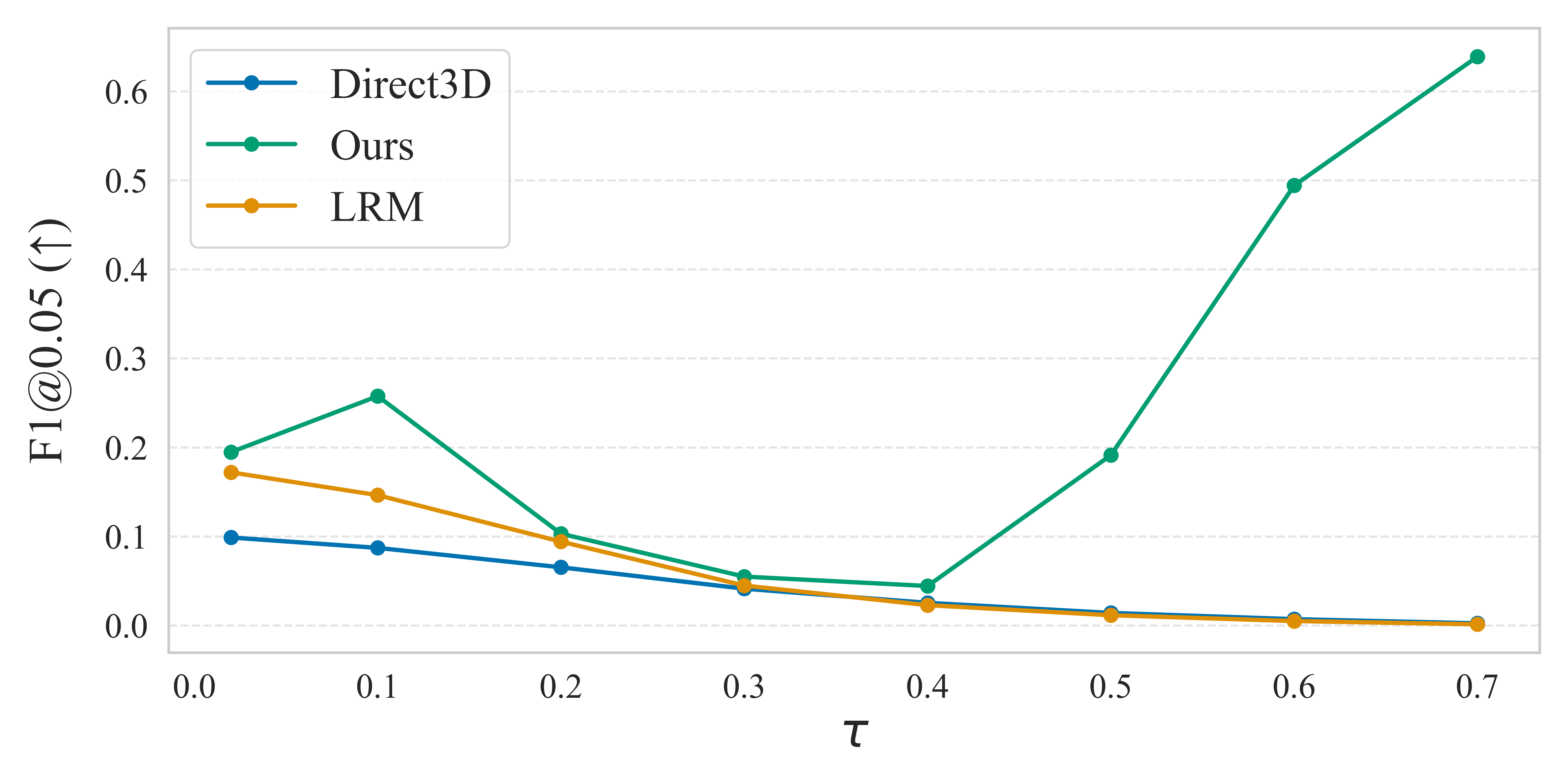}
    \caption{\textbf{F1 score (Higher $\uparrow$ is better)} at distance threshold $0.05$ for unseen object reconstruction as a function of the interaction threshold $\tau$. Our method outperforms both LRM and Direct3D across all $\tau$, indicating improved recovery of surface geometry in multi-level interaction regions.}
    \label{fig:recon_f1}
\end{figure}


\begin{table}[t]
\centering
\caption{Quantitative tactile field reconstruction metrics averaged across all threshold values $\tau$ over $8$ test objects. Higher IoU and F1 scores indicate better volumetric and surface reconstruction, while lower Chamfer Distance indicates better surface reconstruction. We report “-” for Direct3D IoU as their predicted meshes are not view-aligned.}
\begin{tabular}{lccc}
\toprule
\textbf{Method} & \textbf{IoU} $\uparrow$ & \textbf{Chamfer} $\downarrow$ & \textbf{F1@0.05} $\uparrow$ \\
\midrule
\textbf{Ours} & \textbf{0.095} & \textbf{0.2450} & \textbf{0.2474} \\
\textbf{LRM} & 0.052 & 0.3331 & 0.0622 \\
\textbf{Direct3D} & - & 0.3694 & 0.0426 \\
\bottomrule
\end{tabular}
\label{tab:reconstruction_quant_results}
\end{table}

We evaluate planning performance using two metrics: the average planning-length computed using the predicted collision-space $\hat{\mathcal M}_{\tau}$, and the average validity (success rate) of the resulting paths when evaluated against the ground-truth collision-space $\mathcal M_{\tau}$. A path is considered successful if it does not collide with $\mathcal M_{\tau}$, and is considered a failure otherwise. 


\begin{figure*}[!t] 
\centering 
\includegraphics[width=0.9\textwidth]{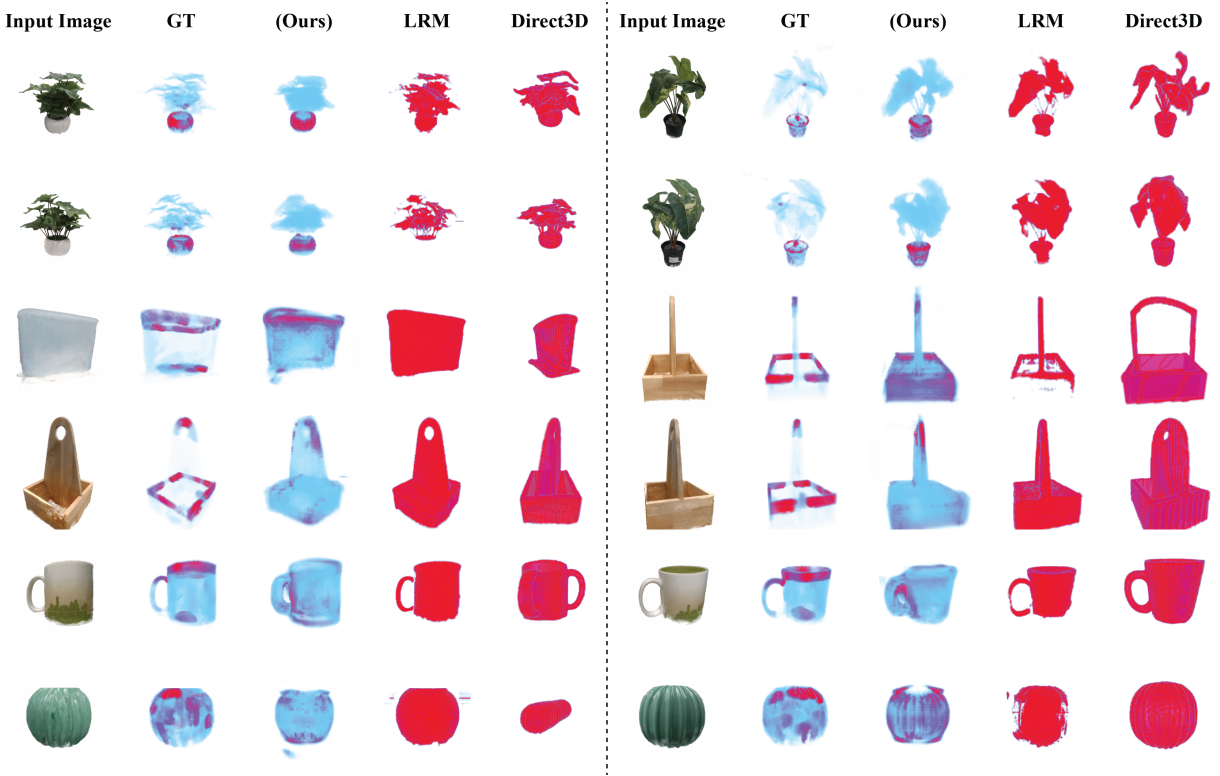} 
\caption{\textbf{Qualitative Reconstruction Results:} Tactile field predictions from monocular RGB images. For each object, we show: the input image, ground-truth (GT) neural tactile field, our prediction, and predictions from LRM and Direct3D baselines. Blue regions indicate low tactile pressure (easily deformable or movable), while red regions indicate high tactile pressure (rigid, heavy, or high-curvature surfaces). Our method accurately predicts that plant foliage is compliant (blue) while pots and bases are resistant (red). Baselines like LRM and Direct3D treat entire objects uniformly, failing to distinguish between deformable and rigid regions. Note that tactile pressure reflects contact in the surface-normal direction; even lightweight objects can exhibit high pressure (red) when pressed against rigid surfaces or structures that resist normal forces, explaining red regions on mug handles and basket edges.} 
\label{fig:reconstruction_qual_results} 
\end{figure*}

\begin{figure*}[!t] \centering \includegraphics[width=\textwidth]{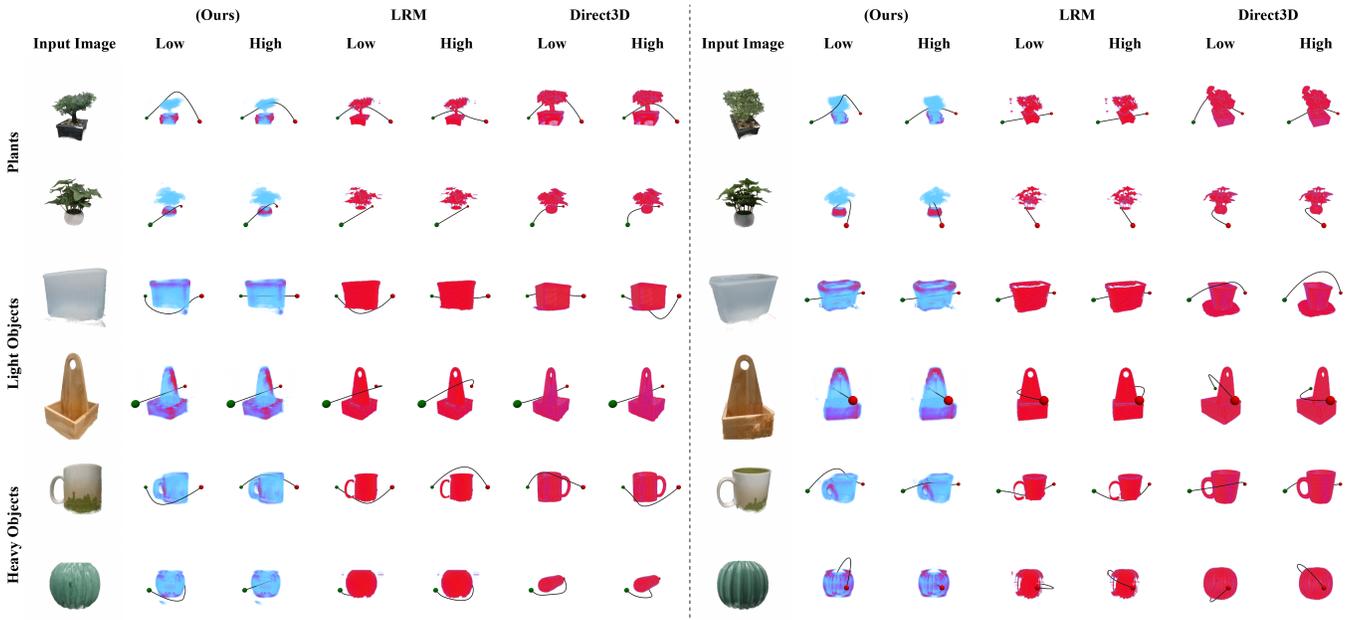} 
\caption{\textbf{Qualitative Planning Results:} Using our predicted tactile field, planned trajectories adapt to the interaction threshold $\tau$. At low thresholds, the planner avoids contact with all objects, treating even lightweight items as obstacles. At high thresholds, the planner embraces contact with compliant objects (e.g., plants, light objects) while avoiding rigid, heavy obstacles (e.g., pots, mugs). Baseline methods using binary occupancy (LRM, Direct3D) treat all objects as equally impassable regardless of threshold, failing to exhibit interaction-aware behavior.} 
\label{fig:planning_qual_results} 
\end{figure*}
To assess benefits of interaction awareness, we compare the average planning length at two interaction thresholds, $\tau = 0.3$ and $\tau = 0.09$. A lower threshold (e.g., $\tau = 0.09$) enforces stricter collision avoidance with the objects, while a higher threshold (e.g., $\tau = 0.3$) allows the planner to route paths through deformable and movable regions of the object. Shorter paths at higher $\tau$ indicate that the predicted tactile field enables the planner to exploit movable or deformable regions to reduce path length.


We compare planning performance against two baselines that use binary occupancy fields: (1) a field constructed from the density predictions of pretrained LRM, and (2) Direct3D, a state-of-the-art 2D image to 3D mesh generative method \cite{wu2024direct3d}. These baselines treat all occupied regions as equally impassable, regardless of physical properties. This comparison demonstrates the advantage of using predicted tactile fields, which encode interaction costs, over geometry-only representations. We note that while prior work on tactile field estimation exists \cite{han2025estimating}, it cannot be directly compared as a baseline because it does not predict tactile fields from monocular images (unlike our method).

\section{Results}
\label{sec:results}

\subsection{Tactile Field Reconstruction}
Our method demonstrates strong performance in 3D reconstruction while providing the additional capability of predicting tactile resistance properties. We evaluate on 8 unseen objects across 120 views per object, comparing against LRM and Direct3D. Figures \ref{fig:recon_chamfer}, \ref{fig:recon_f1} and \ref{fig:recon_iou} show performance across interaction thresholds $\tau$, with average metrics summarized in Table \ref{tab:reconstruction_quant_results}. For a qualitative view of the reconstructed tactile fields, see Figure \ref{fig:reconstruction_qual_results}.

All three metrics (IoU, Chamfer Distance, and F1 score) exhibit systematic variation with $\tau$ due to the ground truth corresponding to progressively smaller regions at higher thresholds. As $\tau$ increases, IoU decreases monotonically for all methods as volumetric overlap shrinks, while Chamfer Distance degrades as surface alignment becomes more challenging. Nevertheless, our method consistently outperforms both baselines across all thresholds and metrics. We achieve an average IoU of 0.095 compared to LRM's 0.052, with nearly twice the IoU at $\tau=0.02$. For surface quality, we attain lower Chamfer Distance and higher F1 scores (Table \ref{tab:reconstruction_quant_results}). Notably, while all methods perform poorly beyond $\tau \geq 0.5$, our approach degrades more gracefully and even improves relative to baselines at higher thresholds. This occurs because baseline methods reconstruct static surfaces regardless of threshold, whereas our method correctly predicts empty sets when no corresponding high-threshold superlevel set exists in the ground truth. Under the metric formulations, this capability directly reduces error by avoiding spurious predictions at thresholds where ground truth surfaces are sparse or absent.

The substantial performance gains over both LRM and Direct3D (as shown in Figures \ref{fig:recon_chamfer}, \ref{fig:recon_f1} and \ref{fig:recon_iou}) arise from our joint optimization of photometric and volumetric objectives. Direct3D optimizes occupancy classification losses \cite{wu2024direct3d}, while LRM optimizes photometric losses \cite{hong2023lrm}. While multi-view photometric supervision encourages view-consistent geometry, it provides weak 3D constraints and often results in degraded reconstruction in occluded regions. Conversely, relying solely on occupancy classification can yield detailed geometry but fails to properly align the reconstructed shape with the input view, as evident in the Direct3D results. By jointly optimizing photometric and volumetric loss terms, our approach balances geometric detail and view alignment, leading to superior performance across all reconstruction metrics.

\subsection{Interaction-Aware Planning}
Beyond reconstruction quality, the predicted tactile fields enable robots to reason about traversable versus non-traversable regions based on contact resistance. We evaluate whether these predictions support more efficient planning by comparing path behavior at different interaction thresholds.

The planned trajectory varies with the threshold parameter $\tau$, resulting in corresponding differences in path length. For a qualitative view of paths discovered over different $\tau$ thresholds, see Figure \ref{fig:planning_qual_results}. At low thresholds (e.g., $\tau=0.09$), our planner produces paths that avoid both plants and heavy objects while allowing traversal through the most lightweight objects (e.g. the wooden organizer). At higher thresholds (e.g., $\tau=0.3$), plans pass through light objects, plant leaves, and even some heavy objects, while still avoiding truly immovable obstacles such as the plant pots. At the lowest thresholds, the resulting behavior closely resembles baseline methods, yielding paths that entirely avoid collisions. However, increasing the threshold allows the planner to traverse regions with higher resistance, enabling shorter overall paths compared to geometry-only baselines.

\begin{table}[t]
\centering
\scriptsize
\setlength{\tabcolsep}{4pt}
\caption{Collision-free path length (len) and success rate (suc) across object categories at low ($\tau=0.09$) and high ($\tau=0.3$) interaction thresholds. Bold entries indicate our method achieves shorter paths by routing through movable/deformable regions while maintaining competitive success rates. All paths are validated against ground-truth tactile fields.}
\begin{tabular}{l cc cc cc cc cc cc}
\toprule
 & \multicolumn{4}{c}{\textbf{Plants}} & \multicolumn{4}{c}{\textbf{Light Objects}} & \multicolumn{4}{c}{\textbf{Heavy Objects}} \\
\cmidrule(lr){2-5} \cmidrule(lr){6-9} \cmidrule(lr){10-13}
 & \multicolumn{2}{c}{Low $\tau$} & \multicolumn{2}{c}{High $\tau$} & \multicolumn{2}{c}{Low $\tau$} & \multicolumn{2}{c}{High $\tau$} & \multicolumn{2}{c}{Low $\tau$} & \multicolumn{2}{c}{High $\tau$} \\
\cmidrule(lr){2-3} \cmidrule(lr){4-5} \cmidrule(lr){6-7} \cmidrule(lr){8-9} \cmidrule(lr){10-11} \cmidrule(lr){12-13}
\textbf{Method} & len & suc & len & suc & len & suc & len & suc & len & suc & len & suc \\
\midrule
\textbf{Ours} & 3.49 & .48 & 3.39 & .27 & \textbf{3.02} & \textbf{.99} & \textbf{3.00} & \textbf{1.0} & 3.78 & .29 & \textbf{3.12} & .39 \\
\textbf{LRM} & \textbf{3.26} & .25 & \textbf{3.20} & .29 & 3.49 & .78 & 3.49 & .82 & \textbf{3.57} & .05 & 3.55 & .30 \\
\textbf{Direct3D} & 3.37 & \textbf{.56} & 3.39 & \textbf{.66} & 3.49 & .74 & 3.49 & .78 & 3.87 & \textbf{.50} & 3.86 & \textbf{.71} \\
\bottomrule
\end{tabular}
\label{tab:planning_quant_results}
\end{table}

Table~\ref{tab:planning_quant_results} summarizes planning performance across object categories. For lightweight objects, our method achieves shorter paths at $\tau=0.3$ while maintaining high success rates, demonstrating effective exploitation of movable regions. Geometry-only baselines cannot distinguish between regions of varying interaction costs, resulting in uniformly conservative paths regardless of threshold. LRM's shorter path lengths for plants result from reconstruction failures that omit rigid pots (see Figure \ref{fig:planning_qual_results}), creating spurious free space that enables invalid shortcuts with correspondingly low success rates.

Our method exhibits notably lower success rates with respect to Direct3D for plants and heavy objects compared to lightweight objects. This occurs because for plants and heavy objects, Direct3D predicts meshes with inflated volumes compared to ground truth. The planner operating over these inflated geometries takes the long way around, resulting in high path lengths and high success rates due to the excessive collision avoidance. Our tactile fields are not over-inflated, producing more accurate geometric predictions. However, this also means our planner generates paths that come closer to actual object boundaries, occasionally resulting in collisions with heavy objects where prediction uncertainty is higher, resulting in a reduced path length but lower success rate.

\section{Conclusion}

We present a supervised approach for inferring tactile fields, dense 3D maps of interaction cost, from a single RGB monocular image. Our method is the first to predict neural tactile fields from a single monocular RGB image. The resulting representation is directly useful for path planning and also provides a strong initialization for multi-view 3D reconstruction methods, such as NeRFs augmented with physical scene attributes. This capability is particularly impactful for applications that require planning in environments with movable or deformable objects (e.g. agricultural robotics).

Despite these advantages, the current dataset is limited to 40 household objects captured on a single mounting surface, which limits generalization. Future work will expand the dataset to include a broader range of objects and support surfaces with varying frictional properties. Additionally, we plan to extend our scalar tactile fields to neural tensor fields that can capture directional interaction properties. These extensions would enable more expressive tactile representations that better capture the richness of real-world physical interactions.






\bibliographystyle{plainnat}
\bibliography{references}

\end{document}